# Exploring Geometric Deep Learning For Precipitation Nowcasting

*Shan Zhao* [1], *Sudipan Saha* [1,4], *Zhitong Xiong* [1], *Niklas Boers* [2,3], *Xiao Xiang Zhu* [1]

Data Science in Earth Observation, Technical University of Munich (TUM), Ottobrunn, Germany [1]
Earth System Modelling, TUM, Ottobrunn, Germany [2]
Potsdam Institute for Climate Impact Research, Potsdam, Germany [3]
Yardi School of Artificial Intelligence, Indian Institute of Technology Delhi, India [4]

## ABSTRACT

Precipitation nowcasting (up to a few hours) remains a challenge due to the highly complex local interactions that need to be captured accurately. Convolutional Neural Networks rely on convolutional kernels convolving with grid data and the extracted features are trapped by limited receptive field, typically expressed in excessively smooth output compared to ground truth. Thus they lack the capacity to model complex spatial relationships among the grids. Geometric deep learning aims to generalize neural network models to non-Euclidean domains. Such models are more flexible in defining nodes and edges and can effectively capture dynamic spatial relationship among geographical grids. Motivated by this, we explore a geometric deep learning-based temporal Graph Convolutional Network (GCN) for precipitation nowcasting. The adjacency matrix that simulates the interactions among grid cells is learned automatically by minimizing the L1 loss between prediction and ground truth pixel value during the training procedure. Then, the spatial relationship is refined by GCN layers while the temporal information is extracted by 1D convolution with various kernel lengths. The neighboring information is fed as auxiliary input layers to improve the final result. We test the model on sequences of radar reflectivity maps over the Trento/Italy area. The results show that GCNs improves the effectiveness of modeling the local details of the cloud profile as well as the prediction accuracy by achieving decreased error measures.

## 1. INTRODUCTION

Earth observation and remote sensing are crucial technologies for monitoring our living environments [1]. Typically based on accurate and high-resolution radar data, precipitation nowcasting aims to estimate the rainfall intensity in the near future (usually up to 2 hours) [2], which is important for water resources management, agriculture, and disaster management. Precise precipitation prediction has remained challenging due to the multi-scale nature of precipitation, with relevant processes with scales ranging from millimeters (droplet formation and cloud microphysics) to thousands of kilometers (large-scale circulation patterns). Numerical Weather Prediction (NWP) integrates discretized versions of the underlying equations of fluid mechanics as initial value problem [3]; although the accuracy has improved greatly in recent decades [4], there remain considerable uncertainties [5] due to the finite grid resolution, errors in parameterizing unresolved processes, and the chaotic dynamics of the atmosphere. Alternatively, data-driven methods provide promising alternatives to process-based models for precipitation nowcasting. Deep learning (DL) models have been shown great potential in capturing the temporal characteristics of rainfall [2], [6]. In particular, deep generative models have recently been shown to outperform existing NWP and other DL (e.g. CNN) benchmarks in nowcasting precipitation [7]. Moreover, the DL and NWP combined methods drive the black-box DL to be more physics-aware and show an improved sensitivity to heavy rains [8, 9, 10].

Radar intensity maps, collected by local radar stations with the high spatial and temporal resolution, are commonly used as the input of precipitation nowcasting models. For data-driven nowcasting based on such data, Ayzel et al. [6] apply a U-shape encoder-decoder segmentation network to generate the output. ConvLSTM and Trajectory GRU (TrajGRU) [2] can learn the location-invariant/variant structures for recurrent connections.

Precipitation nowcasting depends on complex spatial interactions that aren't fully taken into account by the above-mentioned models. Traditional CNNs are only suitable for grided-data assuming translation-invariance. This assumption doesn't hold when it comes to climate data because the cloud pattern usually displays transformation (e.g., rotation) [2]. Geometric deep learning can overcome such limits by dealing with graph structure data [11]. The graph is widely used for geographic applications. For example, climate networks present observations on each grid cell as nodes and compute the similarity between nodes as edges [12, 13, 14, 15]. This has lead to improved predictability and process understanding of extreme rainfall events [16, 17]. Keisler and Ryan [18] use GNNs to generate multi-variant global weather forecasts. Cachay et al. [19] adopted a graph learning layer to automatically study the adjacency matrix to predict El Niño events. Since precipitation patterns exhibit high spatial and temporal variability, we explore geometric DL for precipitation nowcasting. To be more specific, we expect that the relevant dynamical spatial relationships can be more effectively captured by the nodes and edges of a graph convolutional network(GCN). The main contributions of our work include:

- We explore geometric deep networks for the task of precipitation nowcasting. Specifically, we apply a multi-variate time series GCN to handle the complex relationships between different geographical locations on the radar data [20].
- To further enhance the communication between neighboring pixels, we employ feature layers from the geographical proximity area as the augmented inputs to improve the final results.
- We conduct a comprehensive evaluation to assess both pixelwise accuracy and spatial granularity of the prediction.

## 2. METHODS

We formulate the precipitation nowcasting task as a single-step video prediction problem. Given $S$ past observations $X_i$ ($i = t-S+1, t-S+2, ..., t$), $X_i \in R^{C \times W \times H}$ where $W, H$ are the spatial width and height and $C$ is the number of input channels, our task is to predict the most likely frame at a future time step, i.e. $X_{t+T}$. $t$ is the current time step, and $T$ is the horizon, i.e., how far ahead the model predicts the future. Here, we organize each pixel as a node in the graph and randomly initialize its representation. The node similarity is then learned using a small network by minimizing the final prediction loss. The temporal changes are captured by 1D convolutions along the temporal dimension and the adjacency matrix is shared among the frames in the same sequence. Last, the GCN is adopted to refine the graph structure and outputs the final pixel-wise prediction.

### 2.1. Spatial relationship design

For a given frame, to capture the spatial relationship, we treat each grid cell as a node in the graph. Let us assume that the number of nodes in a frame is $N_{node}$. Since Wu et al. [21] validate that the geographical vicinity is not adequate in representing the closeness in the feature space, we adopt the Graph Learning Layer in [21] to let the model learn the adjacency matrix in an automatic and flexible manner. First, the node embedding is randomly initialized. Then, the linear transformation $\theta$ and nonlinear transformation $f_\alpha$ are applied to the initial embedding. The adjacency matrix is learned by a parameterized function $g_\theta$ which considers the difference between the updated node embedding.

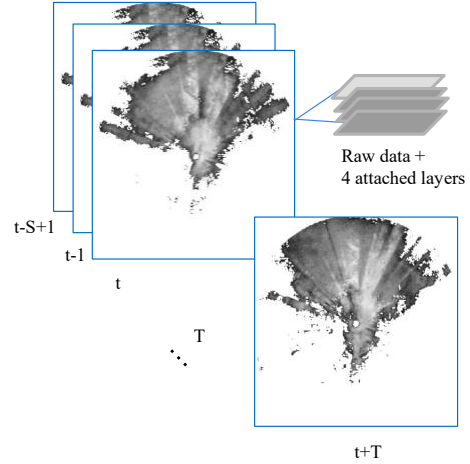

Fig. 1. Single-step precipitation nowcasting given input sequence length $S$ and horizon $T$. The feature dimension is augmented.

$$\mathbf{A} = g_\theta (f_{\alpha 1} (\theta_1 \mathbf{E}_1), f_{\alpha 2} (\theta_2 \mathbf{E}_2)), \quad (1)$$

where $\mathbf{E}_1$ and $\mathbf{E}_2$ are node embeddings. Since in the real world most graphs are sparse and the full matrix contains noisy information, we post-prune the adjacency matrix by only preserving the top $k$ neighbors of each node.

### 2.2. Node representation update

Afterwards, the graph convolutional layer is applied to refine the graph structure and to update the node embedding. The output from this step is an adjacency matrix of size $N_{node} \times N_{node}$ and updated node representation. The same graph structure ($\mathbf{A}$) is shared among different input time steps for fast convergence.

### 2.3. Temporal information extraction

We apply a 1D convolution along the temporal dimension to extract the temporal correlation. Smaller kernels tend to capture the shortterm signal pattern while larger kernels are better at modeling longterm signal pattern. In our method, the inception module [21] is composed of several kernels of various sizes. The temporal features extracted by the differently sized kernels are concatenated with each other to cover all types of movement at different speeds. For one temporal convolutional module, two inception modules [21] run in parallel and are activated by tangent hyperbolic activation and sigmoid activation, respectively, to control the selection and flow of the information. The L1 loss between the prediction and the ground truth observation is computed and minimized.

## 3. EXPERIMENTAL VALIDATION

### 3.1. Dataset

We validate the performance of the model on TAASRAD19 [20]. TAASRAD19 contains radar reflectivity data collected by a singlepolarization Doppler C-Band Radar of a 250 × 250 $km^2$ area of Trentino-Alto Adige/Sudtirol (N 46°29'18", E 11°12'38").

The temporal resolution is 5 minutes and the spatial resolution is 500 meters. To reduce the computation cost, we downsample the image every 5 pixels. To prevent the loss of information caused by downsampling and promote neighboring information extraction, we smooth the raw inputs (without downsampling) using 3×3, 5×5, 9×9, 25×25 mean filter and attach them to the input frame.

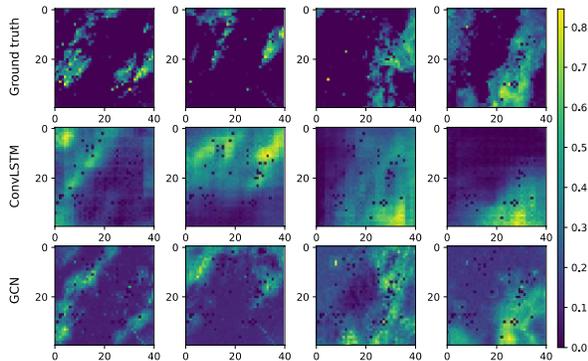

Fig. 2. Results at horizon=10. From top to down are the ground truth image, prediction by ConvLSTM, and prediction by proposed geometric deep learning-based GCN.

All input is normalized to (0,1) as a pre-processing step to make them suitable for the deep learning model. The output is the radar reflectivity which can be converted to rainfall intensity values (mm/h) by the Z-R relationship

$$dBZ = 10\log a + 10b\log R, \quad (2)$$

where $R$ is the rain-rate level, $dBZ$ is the logarithmic radar intensity values, and $a = 58.53$, $b = 1.56$, as provided in [20].

### 3.2. Evaluation

We compute the mean absolute error (MAE), root mean squared error (RMSE), and correlation coefficient (CORR) to evaluate the proximity between the prediction and the ground truth pixel value. To test its functionality in predicting rainfall events, we further computed the Critical Success Index (CSI) and the Heidke Skill Score (HSS) [22]. However, all of the above metrics only measure pixelwise accuracy. Deep learning tends to fool them by producing oversmoothed results [7]. To access the spatial granularity, radiallyaveraged power spectral density (PSD) at different percentiles of all values are reported. All values below 1E-3 are masked during the assessment.

We choose the following single-step prediction baseline models for comparison.

- Mean of input states. The output is simply the average of the input frames.
- ConvLSTM. ConvLSTM is developed from LSTM using convolutional operations as gates. We use a three-layer encoding-forecasting structure with the number of filters for RNNs set to be 64, 192, 192, and kernel sizes equal to 4 × 4, 3 × 3 and 2 × 2.

### 3.3. Results

The network is optimized with the Adam optimizer [23] with the initial learning rate 1e-4 and weight decay 1e-5. We set the batch size to 32. The number of training epochs is 15 for GCN and 20 for ConvLSTM. We set the number of training epochs by inspecting performance on the validation set. The initial node embedding has a dimension of 40 and nonlinearity is activated by *tanh*. The depth of the graph convolutional layer is 2. The top $k$ = 20 neighbors of each node are preserved. We split the data in 2019 year to a ratio of 6/2/2 for training/validation/test set.

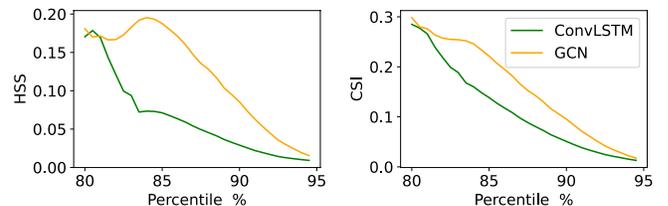

Fig. 3. Metrics to evaluate predictability of rainfall events. CSI and HSS averaged over all test samples are reported.

We evaluate the model performance by comparing it with other temporal prediction methods. The results are shown in Table 1.

We present the accuracy at T=10 as it attains the maximum potential during the nowcasting periods of 1-2 hours. As the horizon increases, the performance of all models decline. GCN outperforms the benchmarks. The graph structure is constantly updated during the learning phase and it has the potential to store temporal information. The mean of the input states performs the worst in the three methods. Figure 2 displays the visual comparison between ground truth and output of deep learning models. Though ConvLSTM wins against GCN by a small margin on correlation, it's shown that convLSTM returns too blurred results due to the large kernel size and simulates fewer details of the cloud shape.

| Method | MAE | RMSE | CORR |
|---|---|---|---|
| Average | 56.8000 | 3.7880 | 0.5231 |
| ConvLSTM [24] | 73.3423 | 3.7261 | **0.6831** |
| Proposed GCN | **55.2068** | **3.2355** | 0.6784 |

Table 1. Error scores. MAE, RMSE, and correlation coefficient are computed at h=10.

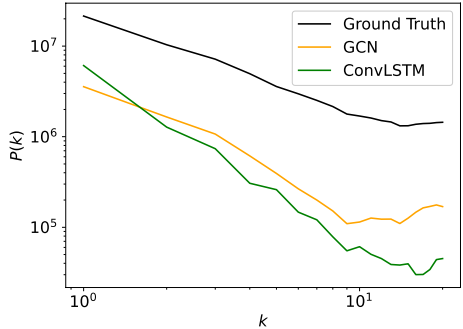

Fig. 4. Metric to evaluate spatial granularity. PSD averaged over all test samples are reported. $k$ is the wavelength number and $P(k)$ is plotted on a double logarithmic scale.

In Figure 3, the CSI and HSS of geometric DL are better than ConvLSTM, indicating overall better prediction of rainfall events. The PSD decreases almost linearly with the increasing wavelength number in Figure 4, i.e., the picture is dominated by coarse features. The geometric DL is slightly better at capturing finer spatial patterns compared with convLSTM.

Further, we test the performance of augmenting feature channels on GCN. The data dimension is shown in Figure 1. The result with and without augmented layers are compared in Table 2. To save time, only 10 sequences are trained for 1 epoch. With the mean filtered layers as the augmented feature dimensions, the error is reduced significantly. However, the training time is increased, which needs to be balanced against the model performance.

| C | MAE | RMSE | CORR | Time(s) |
|---|---|---|---|---|
| 1 | 103.88 | 2.884 | 0.3399 | **85.97** |
| 5 | **68.48** | **2.784** | **0.7603** | 2064.52 |

Table 2. Validation of augmented inputs. C is the number of channels of input.

## 4. CONCLUSION

In this paper, we investigated the potential of geometric DL in the context of precipitation nowcasting on TAASRAD19 radar echo data. Our experiment shows that the geometric DL-based GCN outperforms several baselines and yields improved prediction accuracy including better capturing of details. Especially, PSD validates that the geometric-based method is a promising method to preserve detailed spatial features. Moreover, we validate the merits of using augmented feature layers while dealing with a large number of input pixels. We postulate that the performance will vary with the cloud/precipitation type. In case the divergence brought by different cloud types is large, domain adaptation can be adopted to improve the generalization ability of the model in handling complex weather conditions. In the future, we plan to incorporate other climate variables, such as pressure, wind, and temperature into the model to further improve the prediction accuracy for longer horizons.

## 5. REFERENCES

[1] Zhitong Xiong, Fahong Zhang, Yi Wang, Yilei Shi, and Xiao Xiang Zhu, "Earthnets: Empowering ai in earth observation," *arXiv:2210.04936*, 2022.

[2] Xingjian Shi, Zhihan Gao, Leonard Lausen, Hao Wang, DitYan Yeung, Wai-kin Wong, and Wang-chun Woo, "Deep learning for precipitation nowcasting: A benchmark and a new model," *NeurIPS*, vol. 30, 2017.

[3] Ryuji Kimura, "Numerical weather prediction," *Journal of Wind Engineering and Industrial Aerodynamics*, vol. 90, no. 12-15, pp. 1403–1414, 2002.

[4] Peter Bauer, Alan Thorpe, and Gilbert Brunet, "The quiet revolution of numerical weather prediction," *Nature*, vol. 525, no. 7567, pp. 47–55, 2015.

[5] Azam Moosavi, Vishwas Rao, and Adrian Sandu, "Machine learning based algorithms for uncertainty quantification in numerical weather prediction models," *Journal of Computational Science*, vol. 50, pp. 101295, 2021.

[6] Georgy Ayzel, Tobias Scheffer, and Maik Heistermann, "Rainnet v1. 0: a convolutional neural network for radar-based precipitation nowcasting," *Geoscientific Model Development*, vol. 13, no. 6, pp. 2631–2644, 2020.

[7] Suman Ravuri, Karel Lenc, Matthew Willson, Dmitry Kangin, Remi Lam, Piotr Mirowski, Megan Fitzsimons, Maria Athanassiadou, Sheleem Kashem, Sam Madge, et al., "Skilful precipitation nowcasting using deep generative models of radar," *Nature*, vol. 597, no. 7878, pp. 672–677, 2021.

[8] Philipp Hess and Niklas Boers, "Deep learning for improving numerical weather prediction of heavy rainfall," *Journal of Advances in Modeling Earth Systems*, vol. 14, no. 3, pp. e2021MS002765, 2022.

[9] Christopher Irrgang, Niklas Boers, Maike Sonnewald, Elizabeth A Barnes, Christopher Kadow, Joanna Staneva, and Jan Saynisch-Wagner, "Towards neural earth system modelling by integrating artificial intelligence in earth system science," *Nature Machine Intelligence*, vol. 3, no. 8, pp. 667–674, 2021.

[10] Philipp Hess, Markus Druke, Stefan Petri, Felix M Strnad, and Niklas Boers, "Physically constrained generative adversarial networks for improving precipitation fields from earth system models," *Nature Machine Intelligence*, vol. 4, no. 10, pp. 828–839, 2022.

[11] Sudipan Saha, Shan Zhao, and Xiao Xiang Zhu, "Multitarget domain adaptation for remote sensing classification using graph neural network," *IEEE Geoscience and Remote Sensing Letters*, vol. 19, pp. 1–5, 2022.

[12] Anastasios A Tsonis and Paul J Roebber, "The architecture of the climate network," *Physica A: Statistical Mechanics and its Applications*, vol. 333, pp. 497–504, 2004.

[13] Shraddha Gupta, Niklas Boers, Florian Pappenberger, and Jurgen Kurths, "Complex network approach for detecting tropical cyclones," *Climate Dynamics*, vol. 57, no. 11, pp. 3355–3364, 2021.

[14] Karsten Steinhaeuser, Nitesh V Chawla, and Auroop R Ganguly, "An exploration of climate data using complex networks," in *The Third International Workshop on Knowledge Discovery from Sensor Data*, 2009, pp. 23–31.

[15] Maximilian Gelbrecht, Niklas Boers, and Jurgen Kurths, "A complex network representation of wind flows," *Chaos: An Interdisciplinary Journal of Nonlinear Science*, vol. 27, no. 3, pp. 035808, 2017.

[16] Niklas Boers, Bodo Bookhagen, Henrique MJ Barbosa, Norbert Marwan, Jurgen Kurths, and JA Marengo, "Prediction of extreme floods in the eastern central andes based on a complex networks approach," *Nature communications*, vol. 5, no. 1, pp. 1–7, 2014.

[17] Niklas Boers, Bedartha Goswami, Aljoscha Rheinwalt, Bodo Bookhagen, Brian Hoskins, and Jurgen Kurths, "Complex networks reveal global pattern of extreme-rainfall teleconnections," *Nature*, vol. 566, no. 7744, pp. 373–377, 2019.

[18] Ryan Keisler, "Forecasting global weather with graph neural networks," *arXiv preprint arXiv:2202.07575*, 2022.

[19] Salva Ruhling Cachay, Emma Erickson, Arthur Fender C Bucker, Ernest Pokropek, Willa Potosnak, Salomey Osei, and Bjorn Lütjens, "Graph neural networks for improved el niño forecasting," *arXiv preprint arXiv:2012.01598*, 2020.

[20] Gabriele Franch, Valerio Maggio, Luca Coviello, Marta Pendesini, Giuseppe Jurman, and Cesare Furlanello, "Taasrad19, a high-resolution weather radar reflectivity dataset for precipitation nowcasting," *Scientific Data*, vol. 7, no. 1, pp. 1–13, 2020.

[21] Zonghan Wu, Shirui Pan, Guodong Long, Jing Jiang, Xiaojun Chang, and Chengqi Zhang, "Connecting the dots: Multivariate time series forecasting with graph neural networks," in *ACM SIGKDD international conference on knowledge discovery & data mining*, 2020, pp. 753–763.

[22] Robin J Hogan, Christopher AT Ferro, Ian T Jolliffe, and David B Stephenson, "Equitability revisited: Why the "equitable threat score" is not equitable," *Weather and Forecasting*, vol. 25, no. 2, pp. 710–726, 2010.

[23] Diederik P Kingma and Jimmy Ba, "Adam: A method for stochastic optimization," *arXiv preprint arXiv:1412.6980*, 2014.

[24] Xingjian Shi, Zhourong Chen, Hao Wang, Dit-Yan Yeung, Wai-Kin Wong, and Wang-chun Woo, "Convolutional lstm network: A machine learning approach for precipitation nowcasting," *NeurIPS*, vol. 28, 2015.